\documentclass[11pt]{article}
\usepackage[utf8]{inputenc}
\usepackage[T1]{fontenc}
\usepackage{fixltx2e}
\usepackage{graphicx}
\usepackage{longtable}
\usepackage{float}
\usepackage{wrapfig}
\usepackage{rotating}
\usepackage[normalem]{ulem}
\usepackage{amsmath}
\usepackage{textcomp}
\usepackage{marvosym}
\usepackage{wasysym}
\usepackage{amssymb}
\usepackage{hyperref}
\usepackage[margin=1in]{geometry}
\usepackage{setspace}
\usepackage{glossaries}
\usepackage[export]{adjustbox}
\usepackage{array}
\hypersetup{colorlinks = true, linkcolor = black}
\makeglossaries
\newacronym{relu}{ReLU}{rectified linear unit}
\newacronym{cnn}{CNN}{convolutional neural network}
\newacronym{sgd}{SGD}{stochastic gradient descent}
\newacronym{gpu}{GPU}{graphics processing unit}
\newacronym{wrn}{WRN}{wide residual network}
\author{Robert Kosar \\ David W. Scott}
\date{\today}
\title{The Hybrid Bootstrap: A Drop-in Replacement for Dropout}
\hypersetup{
  pdfkeywords={},
  pdfsubject={},
  pdfcreator={Emacs 24.4.1 (Org mode 8.2.10)}}
\begin{document}

\maketitle
\begin{abstract}
Regularization is an important component of predictive model building.
The hybrid bootstrap is a regularization technique that functions
similarly to dropout except that features are resampled from other
training points rather than replaced with zeros.  We show that the
hybrid bootstrap offers superior performance to dropout.  We also
present a sampling based technique to simplify hyperparameter choice.
Next, we provide an alternative sampling technique for convolutional
neural networks.  Finally, we demonstrate the efficacy of the hybrid
bootstrap on non-image tasks using tree-based models.  The code used
to generate this paper is available
\href{http://github.com/r-kosar/hybrid_bootstrap}{here}.
\end{abstract}

\section{Introduction}
\label{sec-1}
\label{introduction} The field of machine learning offers many potent
models for inference.  Unfortunately, simply optimizing how well these
models perform on a fixed training sample often leads to relatively
poor performance on new test data compared to models that fit the
training data less well.  Regularization schemes are used to constrain
the fitted model to improve performance on new data.

One popular regularization tactic is to corrupt the training data with
independently sampled noise.  This constrains the model to work on
data that is different from the original training data in a way that
does not change the correct inference.  Sietsma and Dow demonstrated
that adding Gaussian noise to the inputs improved the generalization
of neural networks \cite{sietsma1991creating}.  More recently,
Srivastava et al. showed that setting a random collection of layer
inputs of a neural network to zero for each training example greatly
improved model test performance \cite{srivastava2014dropout}.

Both of these types of stochastic regularization have been shown to be
roughly interpretable as types of weight penalization, similar to
traditional statistical shrinkage techniques.  Bishop showed that
using a small amount of additive noise is approximately a form of
generalized Tikhonov regularization \cite{bishop1995training}.  Van der
Maaten et al. showed that dropout and several other types of sampled
noise can be replaced in linear models with modified loss functions
that have the same effect \cite{van2013learning}. Similarly, Wager et
al. showed that, for generalized linear models, using dropout is
approximately equivalent to an $l^2$ penalty following a scaling of the
design matrix \cite{wager2013dropout}.

As noted by Goodfellow et al., corrupting noise can be viewed as a
form of dataset augmentation \cite{goodfellow2016deep}.  Traditional
data augmentation seeks to transform training points in ways that may
drastically alter the point but minimally change the correct
inference.  Corruption generally makes the correct inference more
ambiguous.  Often, effective data augmentation requires domain-specific
knowledge.  However, data augmentation also tends to be much more
effective than corruption, presumably because it prepares models for
data similar to that which they may actually encounter.  For example,
DropConnect is a stochastic corruption method that is similar to
dropout except that it randomly sets neural network weights, rather
than inputs, to zero.  Wan et al. showed that DropConnect (and dropout)
could be used to reduce the error of a neural network on the MNIST
\cite{lecun1998gradient} digits benchmark by roughly 20 percent.
However, using only traditional augmentation they were able to reduce
the error by roughly 70 percent \cite{wan2013regularization}.  Since
corruption seems to be a less effective regularizer than traditional
data augmentation, we improved dropout by modifying it to be more
closely related to the underlying data generation process.

\subsection{The Hybrid Bootstrap}
\label{sec-1-1}
   \label{thehb} 
\begin{figure}[htb]
\centering
\includegraphics[width=.9\linewidth]{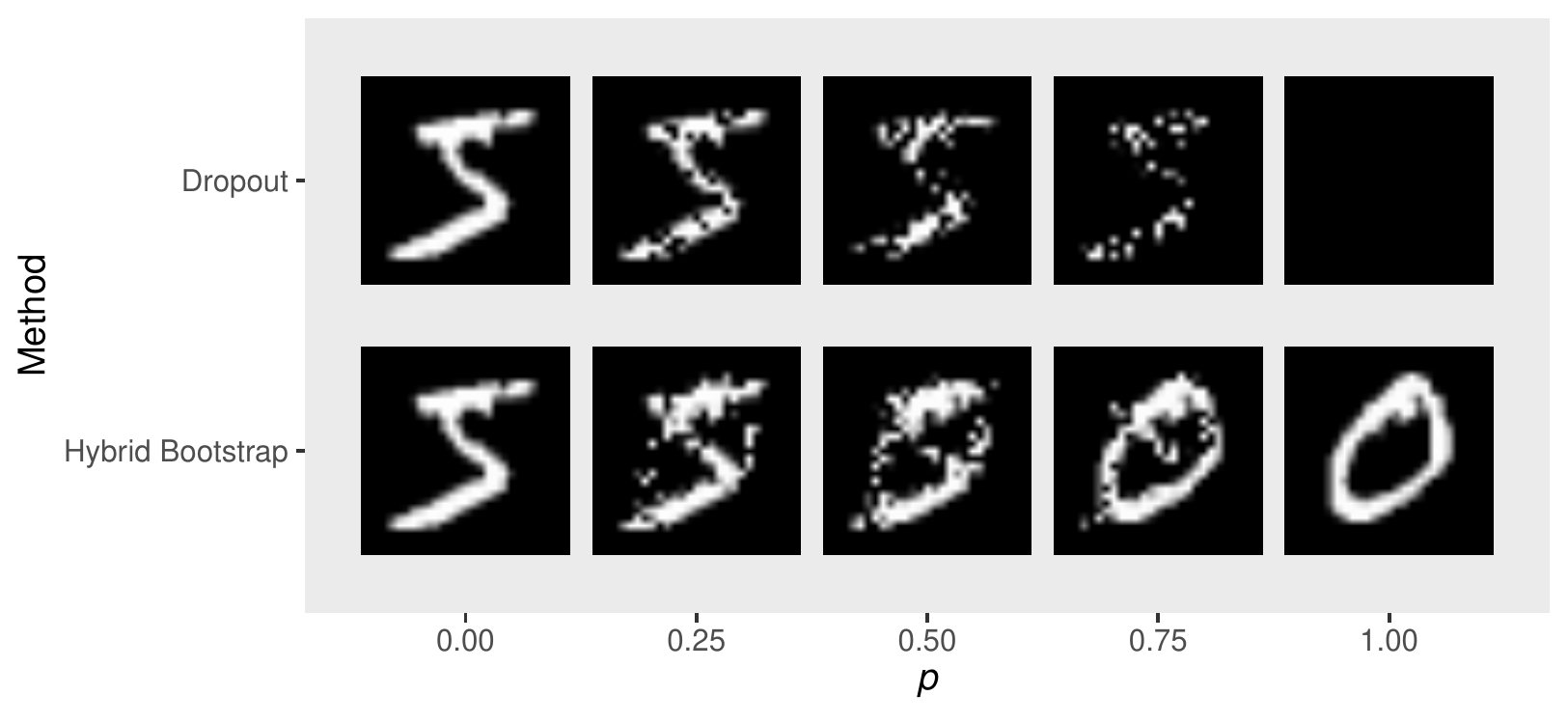}
\caption{\label{hyb_drop_visual}Dropout randomly (with probability $p$) sets a selection of covariates to zero. The hybrid bootstrap randomly replaces a selection of covariates with those from another training point.  The dropout samples in this figure have not been normalized for their corruption levels as would typically be done when used for training.}
\end{figure}

An obvious criticism of dropout as a data augmentation
scheme is that one does not usually expect to encounter randomly
zeroed features in real data, except perhaps in the ``nightmare at test
time'' \cite{globerson2006nightmare} scenario where important features
are actually anticipated to be missing.  One therefore may wish to
replace some of the elements of a training point with values more
plausible than zeros.  A natural solution is to sample a replacement
from the other training points.  This guarantees that the replacement
arises from the correct joint distribution of the elements being
replaced.  We call this scheme the hybrid bootstrap because it
produces hybrids of the training points by bootstrap
\cite{efron1994introduction} sampling.  More formally, define
$\boldsymbol{x}$ to be a vectorized training point.  Then a dropout
sample point $\tilde{\boldsymbol{x}}$ of $\boldsymbol{x}$ is
\begin{equation}
\label{dropout_def}
\tilde{\boldsymbol{x}} = \frac{1}{1-p}\boldsymbol{x} \odot \boldsymbol{\epsilon},
\end{equation}
where $\odot$ is the elementwise product, $\boldsymbol{\epsilon}$ is a
random vector of appropriate dimension such that $\epsilon_i \sim
Ber(1 - p)$, and $p \in [0,1]$.  The normalization by $\frac{1}{1-p}$
seems to be in common use, although it was not part of the original
dropout specification \cite{srivastava2014dropout}.  We then define a
hybrid bootstrap sample point
$\boldsymbol{\overset{\overset{\large.}{\cup}}{x}}$ as
\begin{equation}
\label{hb_def}
\boldsymbol{\overset{\overset{\large.}{\cup}}{x}} =  \boldsymbol{x} \odot \boldsymbol{\epsilon} 
+  \boldsymbol{\overset{\large.}{\cup}} \odot (\boldsymbol{1} - \boldsymbol{\epsilon}),
\end{equation}
where $\boldsymbol{\overset{\large.}{\cup}}$ is a vectorized random training
point and $\boldsymbol{1}$ is a vector of ones of appropriate
length.  In Figure \ref{hyb_drop_visual}, we compare dropout and hybrid
bootstrap sample points for a digit 5.

Typically dropout is performed with the normalization given in
Equation \ref{dropout_def}, but we do not use that normalization for this
figure because it would make the lightly corrupted images dim; we do
use the normalization elsewhere for dropout.  This normalization does
not seem to be useful for the hybrid bootstrap.  One clear difference
between the hybrid bootstrap and dropout for the image data of Figure
\ref{hyb_drop_visual} is that the dropout corrupted sample point remains
recognizable even for corruption levels greater than 0.5, whereas the
hybrid bootstrap sample, unsurprisingly, appears to be more strongly
the corrupting digit 0 at such levels.  In general, we find that lower
fractions of covariates should be resampled for the hybrid bootstrap
than should be dropped in dropout.

\subsection{Paper outline}
\label{sec-1-2}
\label{outline} In this paper, we focus on applying the hybrid
bootstrap to image classification using \glspl{cnn}
\cite{lecun1989backpropagation} in the same layerwise way dropout is
typically incorporated. The basic hybrid bootstrap is an effective
tool in its own right, but we have also developed several
refinements that improve its performance both for general
prediction purposes and particularly for image classification.  In
Section \ref{choose_p}, we discuss a technique for simplifying the choice
of the hyperparameter $p$ for the hybrid bootstrap and dropout.  In
Section \ref{conv_sampling}, we introduce a sampling modification that
improves the performance of the hybrid bootstrap when used with
convolutional neural networks.  In Section \ref{perf_vs_size}, we compare
the performance of the hybrid bootstrap and dropout for different
amounts of training data.  Section \ref{benchmarks} contains results on
several standard benchmark image datasets.  The hybrid bootstrap is
a useful addition to models besides convolutional neural networks;
we present some performance results for the multilayer perceptron
\cite{rosenblatt1961principles} and gradient boosted trees
\cite{friedman2001greedy} in Section \ref{other_algorithms}.

\section{CNN Implementation Details}
\label{sec-2}
\label{implementation} 
\begin{figure}[htb]
\centering
\includegraphics[height=3.5in]{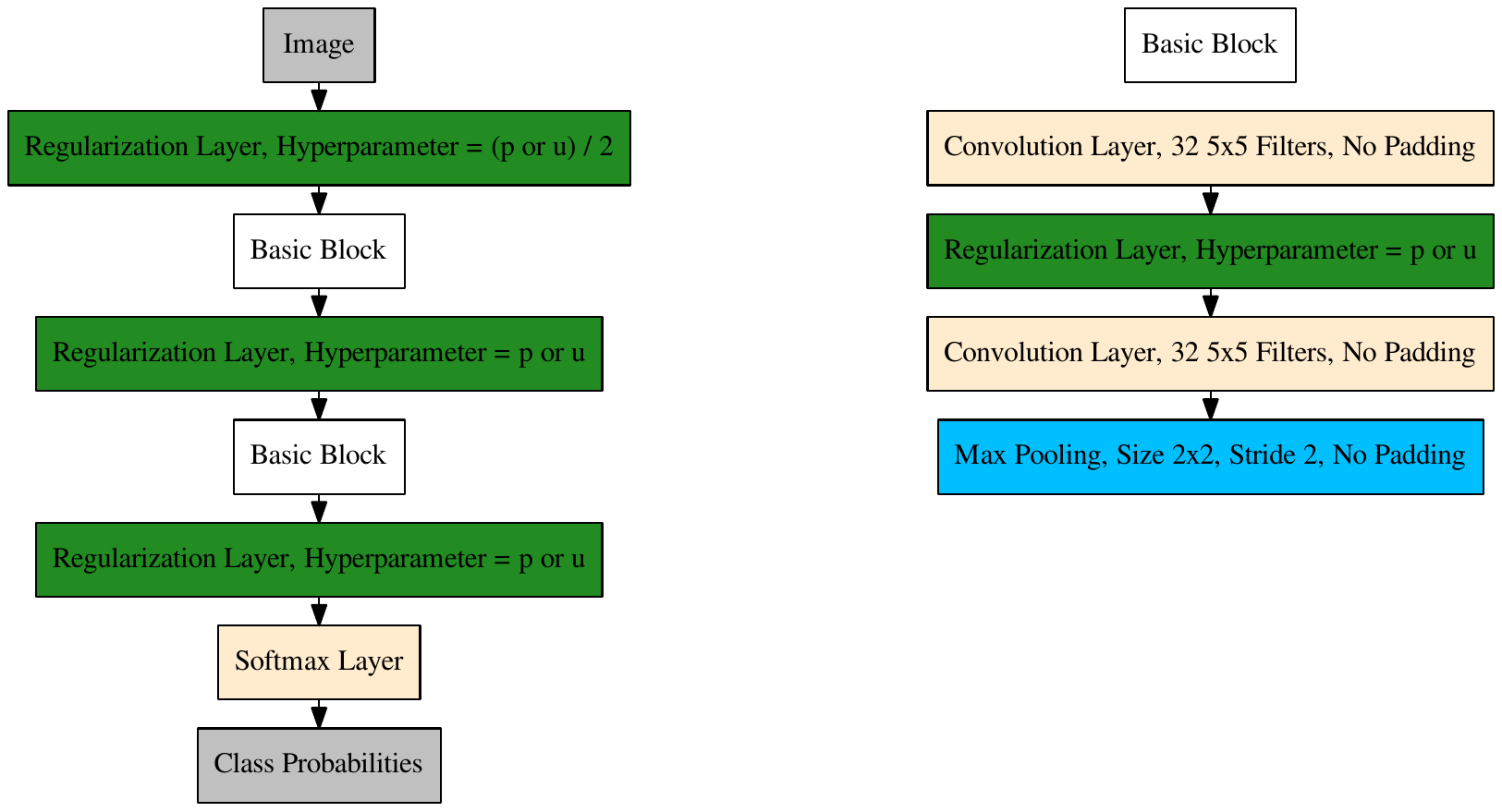}
\caption{\label{experiment_architecture}Network architecture for \gls{cnn} experiments in this paper (except the benchmark results).}
\end{figure}

We fit the \glspl{cnn} in this paper using
backpropagation and \gls{sgd} with momentum.  The hybrid bootstrap
requires selection of $\boldsymbol{\overset{\large.}{\cup}}$ in
Equation \ref{hb_def} for each training example.  We find that using a
$\boldsymbol{\overset{\large.}{\cup}}$ corresponding to the same
training point to regularize every layer of a neural network leads
to worse performance than using different points for each layer.
We use a shifted version of each training minibatch to serve as the
collection of $\boldsymbol{\overset{\large.}{\cup}}$ required to
compute the input to each layer.  We increment the shift by one for
each layer. In other words, the corrupting noise for the first
element of a minibatch will come from the second element for the
input layer, from the third element for the second layer, and so on.
We use a simple network architecture given in
Figure \ref{experiment_architecture} for all \gls{cnn} examples except
those in Section \ref{benchmarks}.  All activation functions except for
the last layer are \glspl{relu} \cite{nair2010rectified}.  As Srivastava
et al. \cite{srivastava2014dropout} found for dropout, we find that
using large amounts of momentum is helpful for obtaining peak
performance with the hybrid bootstrap and generally use a momentum
of 0.99 by the end of our training schedules, except in
Section \ref{benchmarks}.  We make extensive use of both Keras
\cite{chollet2015keras}, and Theano \cite{2016arXiv160502688short} to
implement neural networks and train them on \glspl{gpu}.

\section{Choosing $p$}
\label{sec-3}
\label{choose_p}
\begin{figure}[htb]
\centering
\includegraphics[width=.9\linewidth]{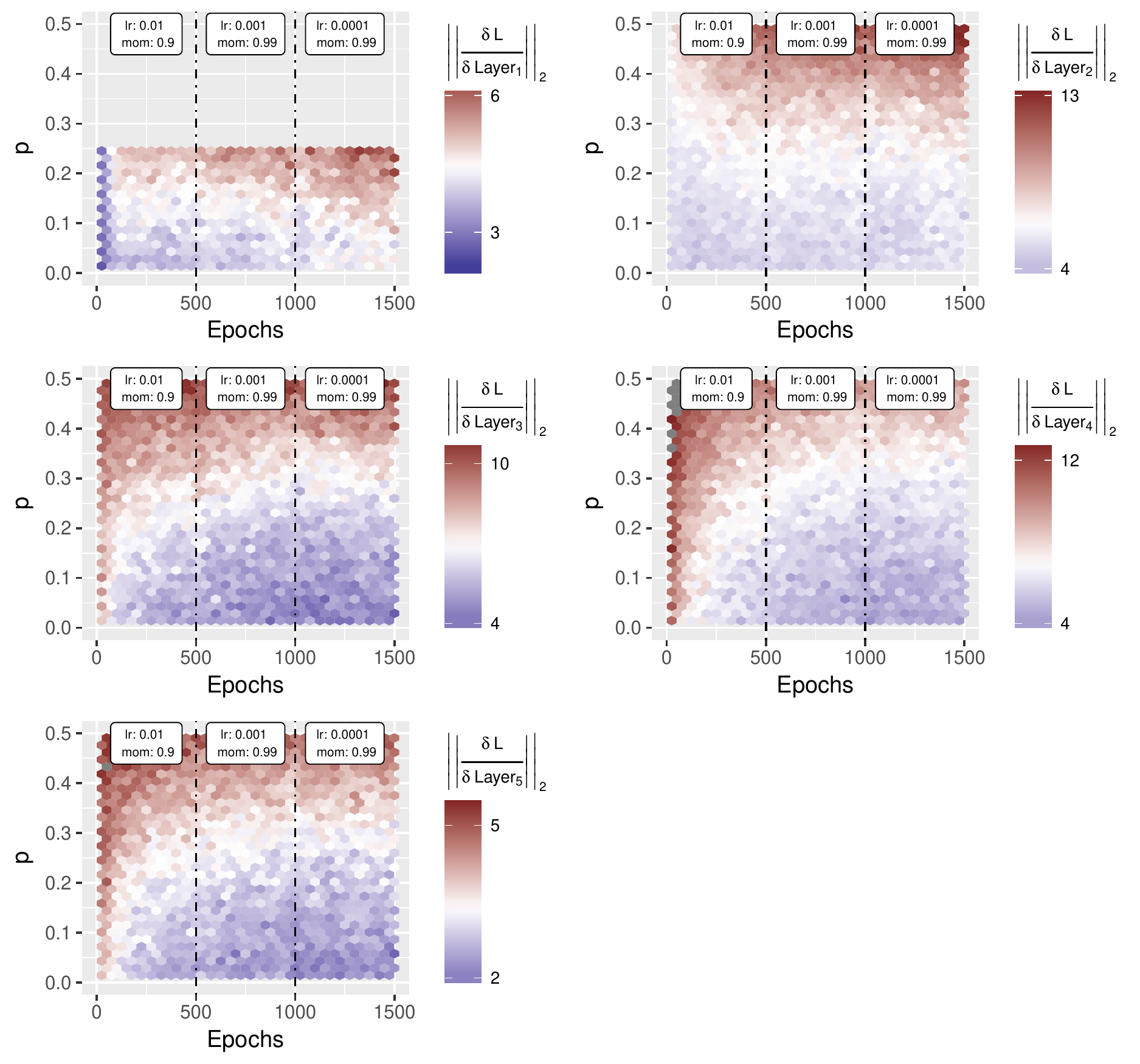}
\caption{\label{gradient_figure}As training progresses, training points that have been corrupted less have smaller gradients than more heavily corrupted points.}
\end{figure}

The basic hybrid bootstrap requires selection of a hyperparameter $p$
to determine what fraction of inputs are to be resampled.  This is a
nuisance because the quality of the selection can have a dramatic
effect on performance, and a lot of computational resources are
required to perform cross validation for complicated networks.
Fortunately, $p$ need not be a fixed value, and we find that sampling
$p$ for each hybrid bootstrap sample to be effective.  We sample from
a $\text{Uniform}(0, u)$ distribution.  Sampling in this way offers
two advantages relative to using a single value:
\begin{enumerate}
\item Performance is much less sensitive to the choice of $u$ than it is
to the choice of $p$ (i.e. tuning is easier).
\item Occasionally employing near-zero levels of corruption ensures
that the model performs well on the real training data.
\end{enumerate}
\begin{figure}[htb]
\centering
\includegraphics[width=.9\linewidth]{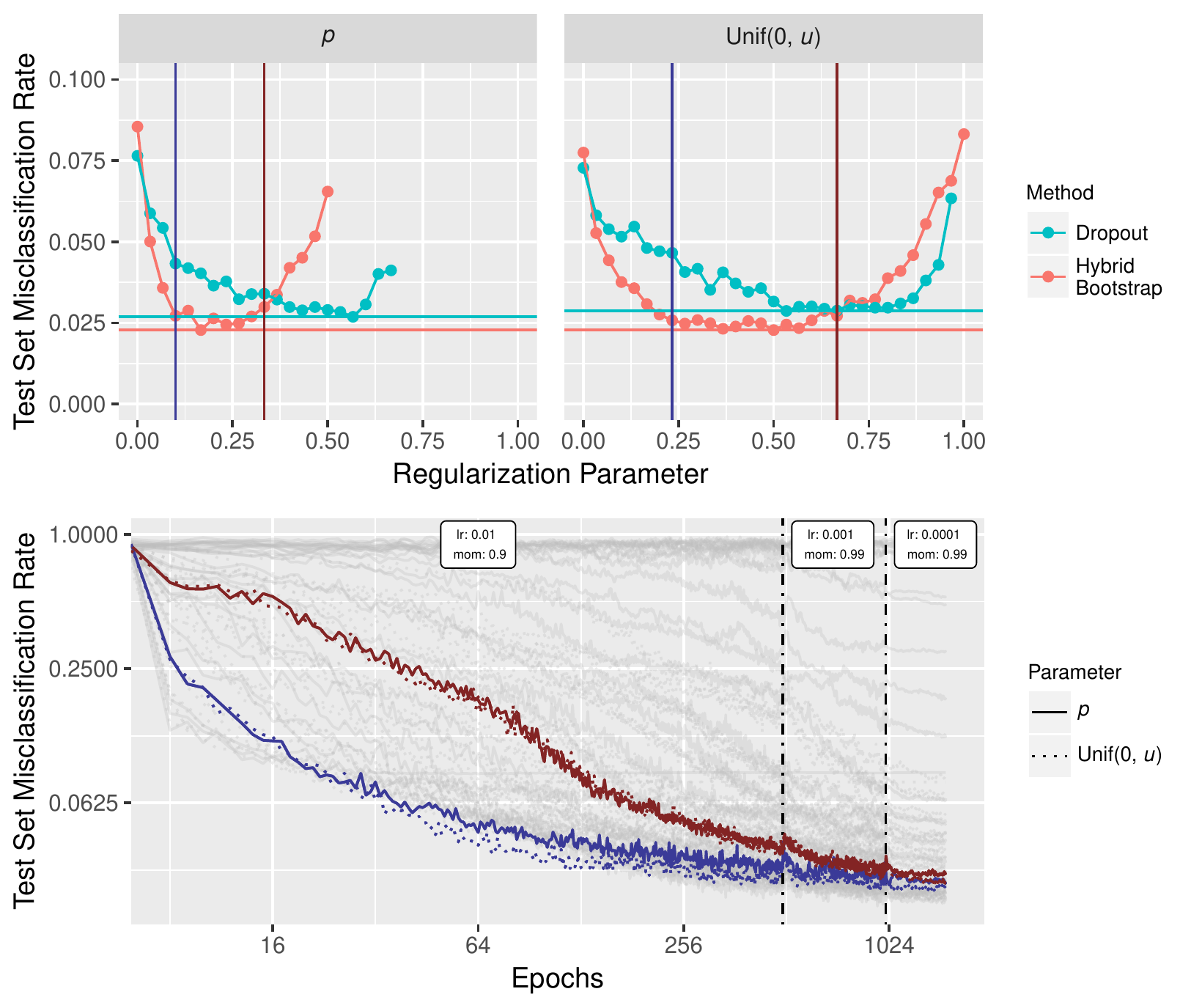}
\caption{\label{loss_vs_p}Test set performance (top) and test set performance over the course of training (bottom) using a  constant hyperparameter vs. a sampled hyperparameter for dropout and the hybrid bootstrap on the MNIST digits with 1,000 training examples. The colored lines in the bottom panel correspond to the regularization levels indicated in the top panel. Sampled hyperparameters perform as well as constant hyperparameters but are much less sensitive to the choice of $u$ than to $p$.}
\end{figure}
The first advantage is illustrated in the top panel of Figure
\ref{loss_vs_p}.  Clearly there are many satisfactory choices of $u$ for
both the hybrid bootstrap and dropout, whereas only a narrow range of
$p$ is nearly optimal.  However, as the bottom panel of Figure
\ref{loss_vs_p} demonstrates, this insensitivity is somewhat contingent upon
training for a sufficient number of epochs. The advantages of sampling
a regularization level for each training point follow from the way
neural networks are fit.  \gls{sgd} and \gls{sgd} with momentum update the
parameters of a neural network by translating them in the direction of
the negative gradient of a minibatch or a moving average of the
negative gradient of minibatches respectively.  The minibatch gradient
can be written as
\begin{equation}
\label{sgd_gradient}
\frac{1}{m}\nabla_{\boldsymbol{\theta}}\sum_{i = 1}^mL(f(\boldsymbol{x}^{(i)};\boldsymbol{\theta}), \boldsymbol{y}^{(i)}),
\end{equation}
where $m$ is the number of points in the batch, $\boldsymbol{\theta}$ is
the vector of model parameters, $L$ is the loss, $f$ is the model,
$\boldsymbol{x}^{(i)}$ is the ith training example in the batch, and $\boldsymbol{y}^{(i)}$ is
the target of the ith training example in the batch \cite{goodfellow2016deep}. Equation \ref{sgd_gradient}
can be rewritten using the chain rule as
\begin{equation}
\label{chain_sgd_gradient}
\frac{1}{m}\sum_{i = 1}^m \left[ \nabla_{f(\boldsymbol{x}^{(i)};\boldsymbol{\theta})}
L(f(\boldsymbol{x}^{(i)};\boldsymbol{\theta}), \boldsymbol{y}^{(i)}) \cdot 
\frac{Df(\boldsymbol{x}^{(i)};\boldsymbol{\theta})}{d\boldsymbol{\theta}} \right].
\end{equation}
The gradient of the loss in Equation \ref{chain_sgd_gradient} is ``small''
when the loss is small; therefore, the individual contribution to the
minibatch gradient is small from individual training examples with
small losses.  As training progresses, the model tends to have
relatively small losses for relatively less-corrupted training points.
Therefore, less-corrupted examples contribute less to the gradient
after many epochs of training.  We illustrate this in Figure
\ref{gradient_figure} by observing the Euclidean norm of the gradient in
each layer as training of our experimental architecture on 1,000 MNIST
training digits progresses.
Clearly low probabilities of resampling are associated with smaller
gradients.  This relationship is somewhat less obvious for layers far
from the output because the gradient size is affected by the amount of
corruption between these layers and the output.

We have no reason to suppose that the uniform distribution is optimal
for sampling the hyperparameter $p$.  We employ it because:
\begin{enumerate}
\item We can easily ensure that $p$ is between zero and one.
\item Uniformly distributed random numbers are readily available in most software packages.
\item Using the uniform distribution ensures that values of $p$ near
zero are relatively probable compared to some symmetric,
hump-shaped alternatives.  This is a hedge to ensure regularized
networks do not perform much worse than unregularized networks.
For instance, using the uniform distribution helps assure that the
optimization can ``get started,'' whereas heavily corrupted
networks can sometimes fail to improve at all.
\end{enumerate}
There are other plausible substitutes, such as the Beta distribution,
which we have not investigated.

\section{Structured Sampling for Convolutional Networks}
\label{sec-4}
\label{conv_sampling} 

The hybrid bootstrap of Equation \ref{hb_def} does not account for the
spatial structure exploited by \glspl{cnn}, so we investigated whether
changing the sampling pattern based on this structure would improve
the hybrid bootstrap's performance on image tasks.

\begin{figure}[htb]
\centering
\includegraphics[width=0.9\linewidth]{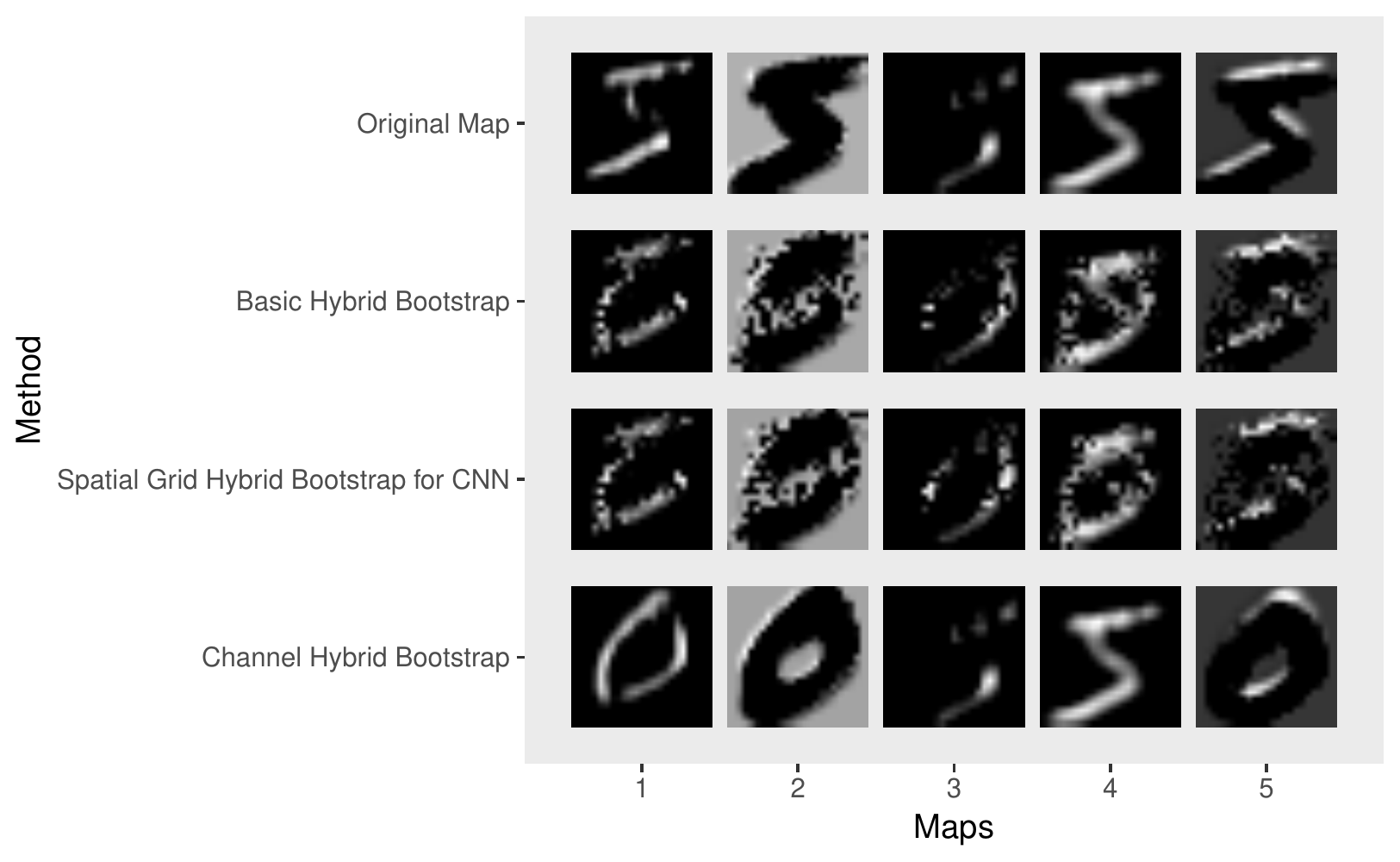}
\caption{\label{sampling_visualization}Visualization of different hybrid bootstrap sampling schemes for \glspl{cnn}.}
\end{figure}

In particular, we wondered if \glspl{cnn} would develop redundant
filters to ``solve'' the problem of the hybrid bootstrap since the
resampling locations are chosen independently for each filter.  We
therefore considered using the same spatial swapping pattern for
every filter, which we call the spatial grid hybrid bootstrap since
pixel positions are either swapped or not.  Tompson et. al
considered dropping whole filters as a modified form of dropout that
they call SpatialDropout (their justification is also spatial)
\cite{tompson2015efficient}.  This approach seems a little extreme in
the case of the hybrid bootstrap because the whole feature map would
be swapped, but perhaps it could work since the majority of feature
maps will still be associated with the target class.  We call this
variant the channel hybrid bootstrap to avoid confusion with the
spatial grid hybrid bootstrap.

The feature maps following regularization corresponding to these
schemes are visualized in Figure \ref{sampling_visualization}. It is
difficult to visually distinguish the spatial grid hybrid bootstrap from the
basic hybrid bootstrap even though the feature maps for the spatial grid
hybrid bootstrap are all swapped at the same locations, whereas the
locations for the basic hybrid bootstrap are independently chosen.
This may explain their similar performance.

\begin{figure}[htb]
\centering
\includegraphics[width=.9\linewidth]{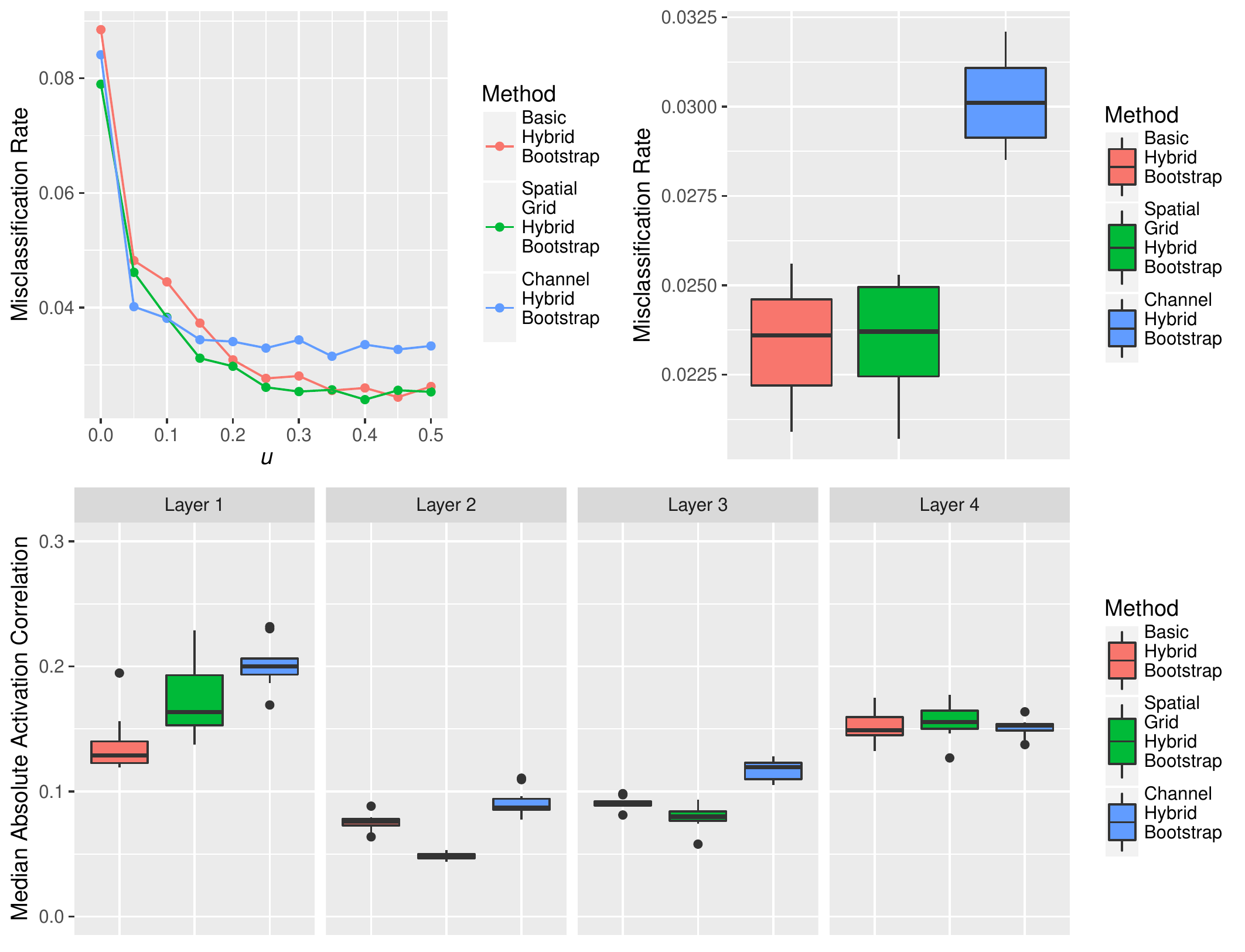}
\caption{\label{sampling_correlation_figure}Validation accuracy of structured sampling schemes using 1,000 training images (top left), test accuracy of structured sampling schemes for 10 different initializations (top right), and median absolute correlations of neurons in convolutional layers following training for 10 initializations (bottom).}
\end{figure}

We compare the error rates of the three hybrid bootstrap schemes in
the top left panel of Figure \ref{sampling_correlation_figure} for various
values of $u$ by training on 1,000 points and computing the accuracy
on the remaining 59,000 points in the conventional training set. Both
the spatial grid and the channel hybrid bootstrap outperform the
basic hybrid bootstrap for low levels of corruption.  As $u$
increases, the basic hybrid bootstrap and the spatial grid hybrid
bootstrap reach similar levels of performance.  Both methods reach a
(satisfyingly flat) peak at approximately $u = 0.45$.  As indicated
in the top right panel of Figure \ref{sampling_correlation_figure}, the
test accuracies of both the basic hybrid bootstrap and the spatial
grid hybrid bootstrap are similar for different initializations of
the network at this chosen parameter. We compare the redundancy of
networks regularized using the three hybrid bootstrap variants at
this level of corruption.

One possible measure of the redundancy of filters in
a particular layer of a \gls{cnn} is the average absolute correlation
between the output of the filters.  We consider the median absolute
correlation for 10 different initializations in the bottom panel of
Figure \ref{sampling_correlation_figure}.  The middle two layers exhibit
the pattern we expected: the spatial grid hybrid bootstrap leads to
relatively small correlations between filters.  However, this
pattern does not hold for the first and last convolutional layer.
If we attempt to reduce the initial absolute correlations of the
filters with a rotation, even this pattern does not hold up.

Overall, the difference in performance between the spatial grid
hybrid bootstrap and the basic hybrid bootstrap is modest,
particularly near their optimal parameter value.  We use the spatial
grid hybrid bootstrap for \glspl{cnn} on the basis that it seems to
perform at least as well as the basic hybrid bootstrap, and
outperforms the basic hybrid bootstrap if we select a $u$ that is
too small.
\section{Performance as a Function of Number of Training Examples}
\label{sec-5}
\label{perf_vs_size}

\begin{figure}[!htpb]
\centering
\includegraphics[width=.9\linewidth]{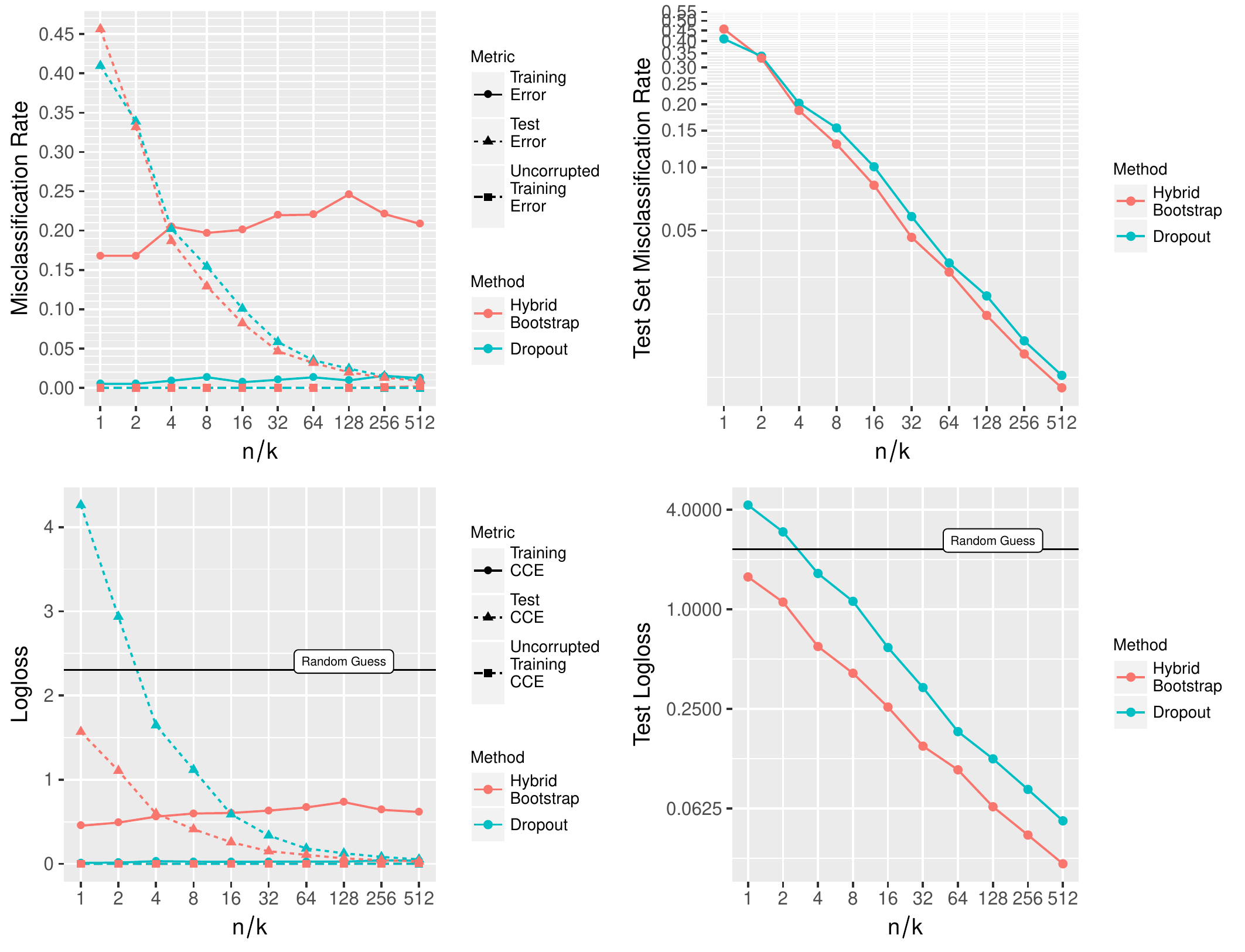}
\caption{\label{perf_vs_size_fig}Performance of the hybrid bootstrap compared to dropout for different training set sizes. Here a random guess assigns a probability of $\frac{1}{k}$ where $k$ is the number of classes.}
\end{figure}

We find the hybrid bootstrap to be particularly effective when only
a small number of training points are available.  In the most
extreme case, only one training point per class exists.  So-called
one-shot learning seeks to discriminate based on a single training
example.  In Figure \ref{perf_vs_size_fig}, we compare the performance of
dropout and the hybrid bootstrap for different training set sizes
using the hyperparameters $u=0.45$ and $u = 0.65$ for the hybrid
bootstrap and dropout respectively.

Both techniques perform remarkably well even for small dataset sizes
but the hybrid bootstrap has a clear advantage.  If one considers
the logloss as a measure of model performance, the hybrid bootstrap
works even when only one or two examples from each class are
available.  However, dropout is less effective than assigning equal
odds to each class for those dataset sizes.  The error rate of the
network on dropout-corrupted data (shown in the top left panel of
Figure \ref{perf_vs_size_fig}) is quite low even though there is a large
amount of dropout.  This comparison is potentially unfair to dropout
as an experienced practitioner may suspect that our test
architecture contains too many parameters for such a small training
set before using it.  However, for the less-experienced who must rely on it, cross
validation is challenging with only one training point.

\section{Benchmarks}
\label{sec-6}
\label{benchmarks} 
\begin{figure}[htb]
\centering
\includegraphics[width=0.9\textwidth]{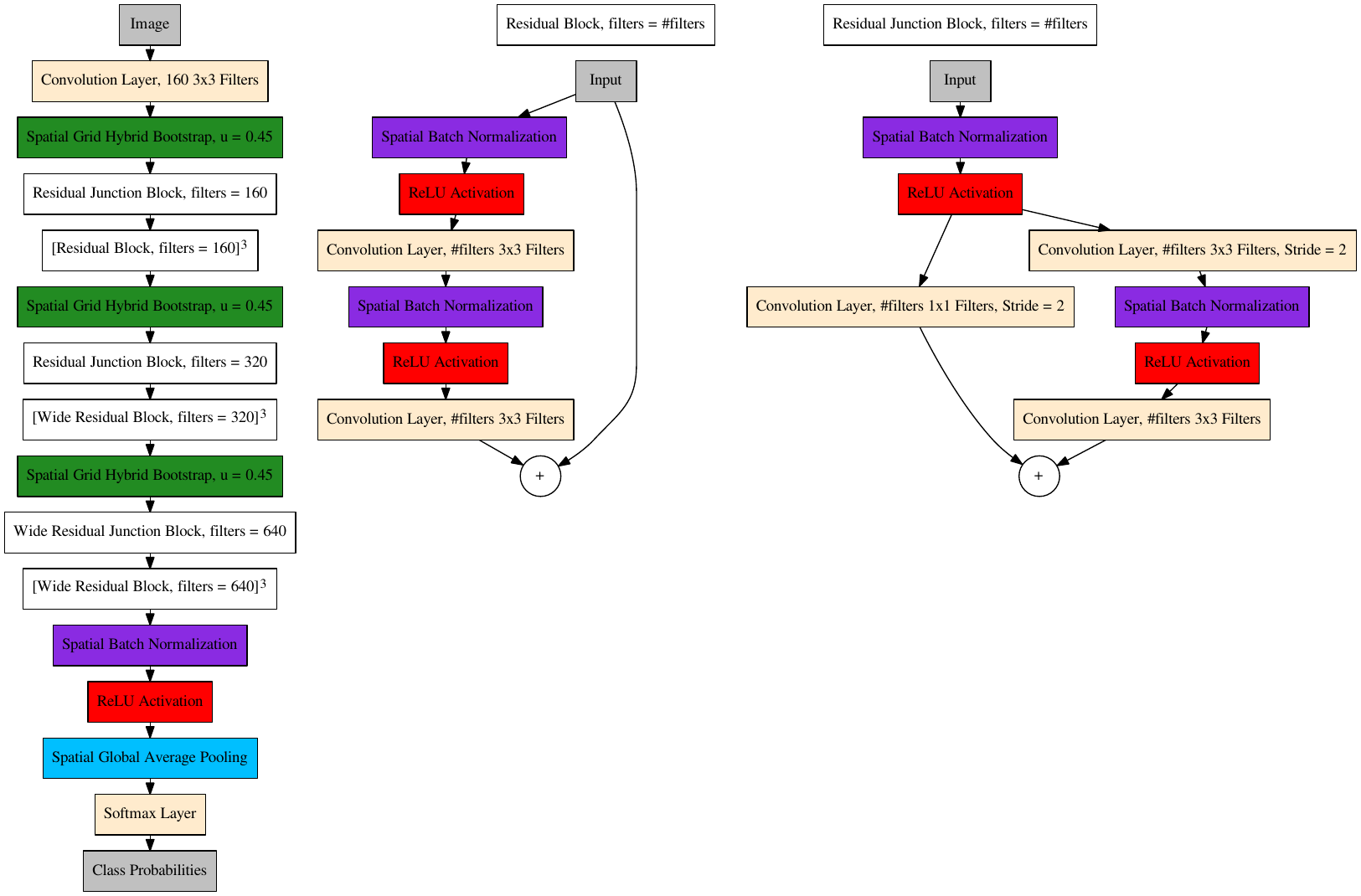}
\caption{\label{benchmark_architecture}Network architecture for benchmark results.  Biases are not used.  All convolutional layers have linear activations. Exponents represent the number of repeated layers.}
\end{figure}

The previous sections employed smaller versions of
the MNIST training digits for the sake of speed, but clearly the
hybrid bootstrap is only useful if it works for larger datasets and
for data besides the MNIST digits.  To evaluate the hybrid
bootstrap's performance on three standard image benchmarks, we adopt
a \gls{cnn} architecture very similar to the \glspl{wrn} of Zagoruyko and
Komodakis \cite{Zagoruyko2016WRN} with three major differences.
First, they applied dropout immediately prior to certain weight
layers.  Since their network uses skip connections, this means
difficult regularization patterns can be bypassed, defeating the
regularization.  We therefore apply the hybrid bootstrap prior to
each set of network blocks at a particular resolution.  Second, we
use 160 rather than 16 filters in the initial convolutional layer.
This allows us to use the same level of hybrid bootstrap for each of
the three regularization layers. Third, their training schedule
halted after decreasing the learning rate three times by 80\%.  Our
version of the network continues to improve significantly at lower
learning rates, so we decrease by the same amount five times. Our
architecture is visualized in Figure \ref{benchmark_architecture}.

We test this network on the CIFAR10 and CIFAR100 datasets, which
consist of RGB images with 50,000 training examples and 10,000 test
cases each and 10 and 100 classes respectively
\cite{krizhevsky2009learning}.  We also evaluate this network on the
MNIST digits. We augment the CIFAR data with 15\% translations and
horizontal flips.  We do not use data augmentation for the MNIST
digits. The images are preprocessed by centering and scaling
according to the channel-wise mean and standard deviation of the
training data.  We use \gls{sgd} with Nesterov momentum 0.9 and start
with learning rate 0.1.  The learning rate is decreased by 80\%
every 60 epochs and the network is trained for 360 epochs total.
The results are given in Table \ref{benchmark_table}.  We attempted to use
dropout in the same position as we use the hybrid bootstrap, but
this worked very poorly.  At dropout levels $p = 0.5$ and $p =
  0.25$, the misclassification rates on the CIFAR100 test set are
50.56\% and
28.83\%
respectively, which is much worse than the hybrid bootstrap result.  To
have a real comparison to dropout, we have included the
dropout-based results from the original wide residual network paper.
It is apparent in Table \ref{benchmark_table} that adding the hybrid
bootstrap in our location makes a much bigger difference than adding
dropout to the residual blocks.

\begin{table}[htb]
\caption{\label{benchmark_table}Benchmark results.  Values are misclassification percentages. The CIFAR datasets are augmented with translations and flips.  The MNIST digits are not augmented.}
\centering
\begin{tabular}{c|cccc}
Dataset & Hybrid Bootstrap & No Stochastic Reg. & Dropout & No Stochastic Reg.\\
 & (Our Architecture) & (Our Architecture) & (\gls{wrn} 28-10) & (\gls{wrn} 28-10)\\
\hline
CIFAR10 & 3.4 & 4.13 & 3.89 & 4.00\\
CIFAR100 & 18.36 & 20.1 & 18.85 & 19.25\\
MNIST & 0.3 & 0.66 & NA & NA\\
\end{tabular}
\end{table}

\section{Other Algorithms}
\label{sec-7}
\label{other_algorithms} The hybrid bootstrap is not only useful for
\glspl{cnn}.  It is also applicable to other inferential algorithms
and can be applied without modifying their underlying code by
expanding the training set in the manner of traditional data
augmentation.
\subsection{Multilayer Perceptron}
\label{sec-7-1}
The multilayer perceptron is not of tremendous modern interest for
image classification, but it is still an effective model for other
tasks.  Dropout is commonly used to regularize the multilayer
perceptron, but the hybrid bootstrap is even more effective.  As an
example, we train a multilayer perceptron on the MNIST digits with
2 ``hidden'' layers of $2^{13}$ neurons each with \gls{relu}
activations and $u = 0.225$, $u = 0.45$ hybrid bootstrap
regularization for each layer respectively.  We use weight decay
0.00001 and \gls{sgd} with momentum 0.9 and batch size 512.  We start
the learning rate at 0.1 and multiply it by 0.2 every 250 epochs
for 1,000 epochs total training.  The resulting network has error
0.81\%
on the MNIST test set.  Srivastava et al. used
dropout on the same architecture with resulting error 0.95\%
\cite{srivastava2014dropout}.  However, their training schedule was
different and they used a max-norm constraint rather than weight
decay.  To verify that this improvement is not simply a consequence
of these differences rather than the result of replacing dropout
with the hybrid bootstrap, we use the same parameters but replace
the hybrid bootstrap with dropout $p = 0.25$, $p = 0.5$
respectively.  The resulting network has test set error
1.06\%.

\subsection{Boosted Trees}
\label{sec-7-2}
One of the most effective classes of prediction algorithms is that
based on gradient boosted trees described by Friedman
\cite{friedman2001greedy}. Boosted tree algorithms are not very
competitive with \glspl{cnn} on image classification problems, but
they are remarkably effective for prediction problems in general
and have the same need for regularization as other nonparametric
models.  We use XGBoost \cite{chen2016xgboost}, a popular
implementation of gradient boosted trees.  

\begin{figure}[htb]
\centering
\includegraphics[width=.9\linewidth]{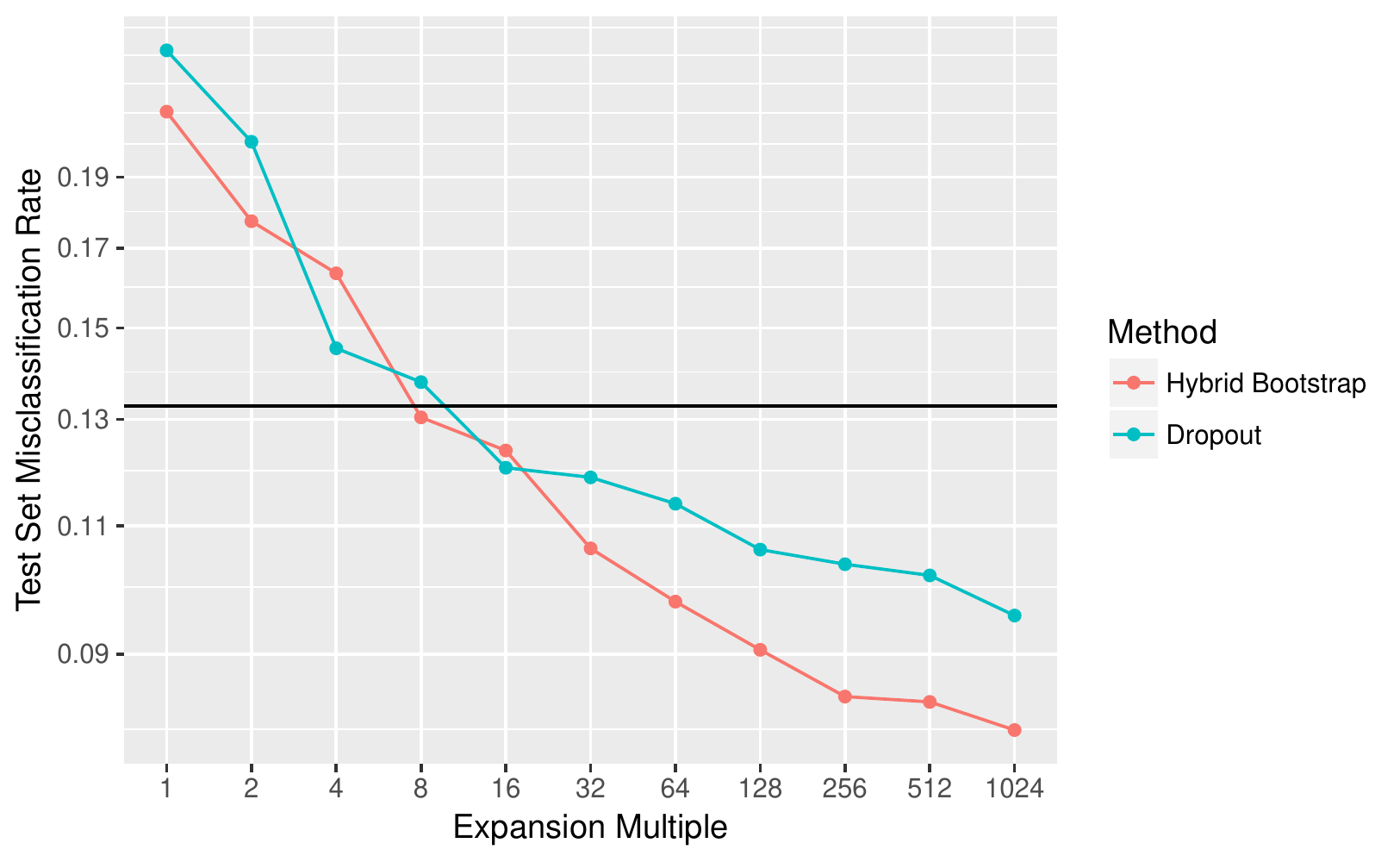}
\caption{\label{mnist_trees}Comparison of boosted tree test set performance on the MNIST digits for stochastic expansions of 1,000 training images. The horizontal line is the performance of an XGBoost model using four times as many trees and a smaller step size, but no additional regularization.}
\end{figure}

Vinayak and Gilad-Bachrach proposed dropping the constituent models
of the booster during training, similar to dropout
\cite{pmlr-v38-korlakaivinayak15}.  This requires modifying the
underlying model fitting, which we have not attempted with the
hybrid bootstrap.  However, if we naively generate hybrid bootstrap
data on 1,000 MNIST digits with the hyperparameters $u = 0.45$ and
$u = 0.65$ for the hybrid bootstrap and dropout respectively, we
can see that the hybrid bootstrap outperforms dropout in Figure
\ref{mnist_trees}.  We note that extreme expansion of the training data
by hybrid bootstrap sampling seems to be important for peak
predictive performance.  However, this must be balanced by
consideration of the computational cost.

\begin{figure}[htb]
\centering
\includegraphics[width=.9\linewidth]{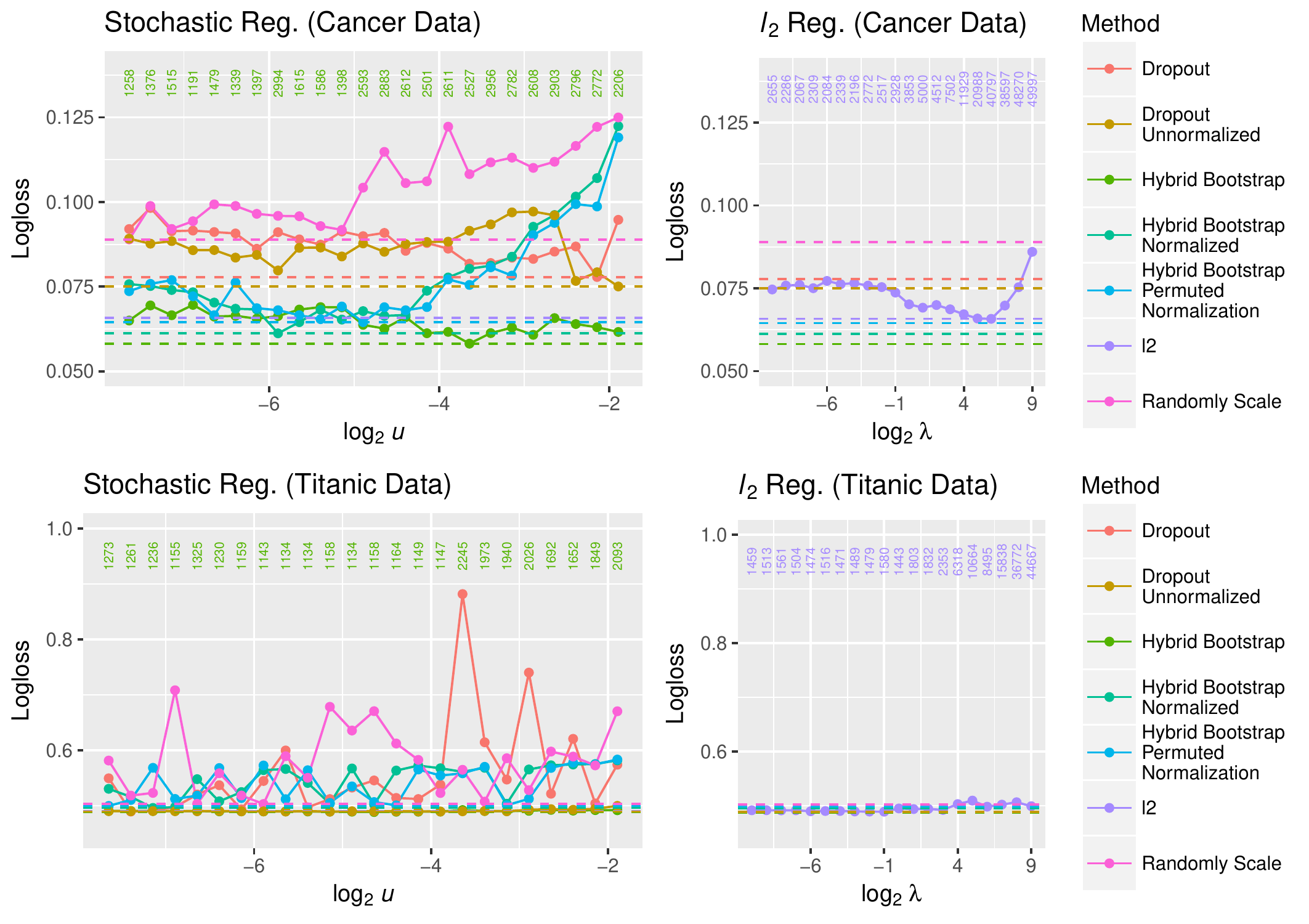}
\caption{\label{xgboost_cancer}Comparison of different regularization mechanisms for boosted trees for breast cancer malignancy (top) and Titanic passenger survival (bottom).  The number of trees selected using cross validation is printed at the top of each panel. Horizontal lines are at the minimum of each curve to aid comparison, but the variation of each curve is of significant importance too.}
\end{figure}

We also compare dropout and the hybrid bootstrap for the breast
cancer dataset where malignancy is predicted from a set of 30
features \cite{street1993nuclear}.  XGBoost provides its own
$l^2$-type regularization technique that we typically set to a
negligible level when using the hybrid bootstrap.  We compare
XGBoost`s $l^2$ regularization with several stochastic methods.
The stochastic methods are: the hybrid bootstrap, dropout, the
hybrid bootstrap with the dropout normalization, the hybrid
bootstrap with a random permutation of the dropout normalization,
dropout without the normalization, and just the dropout
normalization.  We expand the training dataset by a factor of 1,000
for the stochastic methods.  Early experiments on depth-one trees
indicated that simply randomly scaling the data was effective but
this does not seem to apply to other tree depths.  We randomly
split the 569 observations into a training set and a test set and
use the median of a five-fold cross validation to select the
appropriate number of depth-two trees.  We allow at most 50,000
trees for the $l^2$ regularized method and 3,000 trees for the
stochastic regularization methods.  We perform the same the same
procedure for the well-known Titanic survival data \cite{titanic}
except the test set is chosen to be the canonical test set.  We use
only numeric predictors and factor levels with three or fewer
levels for the Titanic data, leaving us with $p = 7$ and 891 and
418 observations in the training and test sets respectively.  The
results are given in Figure \ref{xgboost_cancer}.  The hybrid bootstrap
outperforms the $l^2$ regularization in both cases (very marginally
in the case of the Titanic data).  The normalized stochastic
methods do not seem to be terribly effective, particularly for the
Titanic data.  We suspect this is because they replace the the
factor data with non-factor values.  We note that training using
the augmented data takes much longer than the $l^2$ regularization.
However, the number of trees selected for the $l^2$ regularization
method (printed in the same figure) may be significantly larger
than for the stochastic methods, so the stochastic regularizers may
offer a computational advantage at inference time.  The sizes of
these datasets are small, and we note that the $l^2$ regularization
has a lower (oracle) classification error for the cancer data.  Of
course, nothing prevents one from using both regularization schemes
as part of an ensemble, which works well in our experience.

\section{Discussion}
\label{sec-8}
The hybrid bootstrap is an effective form of regularization.  It can
be applied in the same fashion as the tremendously popular dropout
technique but offers superior performance.  The hybrid bootstrap can
easily be incorporated into other existing algorithms.  Simply
construct hybrid bootstrap data as we do in Section
\ref{other_algorithms}.  Unlike other noising schemes, the hybrid
bootstrap does not change the support of the data.  However, the
hybrid bootstrap does have some disadvantages.  The hybrid bootstrap
requires the choice of at least one additional hyperparameter.  We
have attempted to mitigate this disadvantage by sampling the hybrid
bootstrap level, which makes performance less sensitive to the
hyperparameter.  The hybrid bootstrap performs best when the
original dataset is greatly expanded. The magnitude of this
disadvantage depends on the scenario in which supervised learning is
being used.  We think that any case where dropout is being used is a
good opportunity to use the hybrid bootstrap.  However, there are
some cases, such as linear regression, where the hybrid bootstrap
seems to offer roughly the same predictive performance as existing
methods, such as ridge regression, but at a much higher
computational cost.  The hybrid bootstrap's performance may depend
on the basis in which the data are presented.  This disadvantage is
common to many algorithms.  One reason we think the hybrid bootstrap
works so well for neural networks is that they can create features
in a new basis at each layer that can themselves be hybrid
bootstrapped, so the initial basis is not as important as it may be
for other algorithms.

We have given many examples of the hybrid bootstrap working, but
have devoted little attention to explaining why it works.  There is
a close relationship between hypothesis testing and regularization.
For instance, the limiting behavior of ridge regression is to drive
regression coefficients to zero, a state which is a common null
hypothesis.  The limiting behavior of the hybrid bootstrap is to
make the class (or continuous target) statistically independent of
the regressors, as in a permutation test.  Perhaps the hybrid
bootstrap forces models to possess a weaker dependence between
predictor variables and the quantity being predicted than they
otherwise would.  We recognize this is a vague explanation (and
could be said of other forms of regularization), but we do find that
the hybrid bootstrap has a lot of practical utility.

\section{Miscellaneous Acknowledgments}
\label{sec-9}
While we were writing this paper, Michael Jahrer independently used
the basic hybrid bootstrap as input noise (under the alias ``swap
noise'') for denoising autoencoders as a component of his winning
submission to the Porto Seguro Safe Driver Prediction Kaggle
competition.  Clearly this further establishes the utility of the
hybrid bootstrap!

We have also recently learned that there are currently at least
three distinct groups that have papers at various points in the
publishing process concerning convex combinations of training
points, which are similar to hybrid bootstrap combinations
\cite{convex1,convex2,convex3}.

\printglossaries \bibliographystyle{unsrt}
\bibliography{refs.tex}
\end{document}